%% file: blockgpt_final.tex
\documentclass{article}

\usepackage[final]{tackling_climate_workshop_style}

\usepackage[utf8]{inputenc} 
\usepackage[T1]{fontenc}    

\usepackage{hyperref}       
\usepackage{url}            
\usepackage{booktabs}       
\usepackage{amsfonts}       
\usepackage{xcolor}         
\usepackage{graphicx}
\usepackage{subcaption}
\usepackage{float}
\usepackage{dsfont}
\usepackage{amsmath}
\usepackage{multirow}
\title{BlockGPT: Spatio-Temporal Modelling of Rainfall via Frame-Level Autoregression}

\author{%
  Cristian Meo$^{*}$ \\
  LatentWorlds AI\\
  TUDelft\\
  \texttt{c.meo@tudelft.nl} \\
  Netherlands \\
  \And
  Varun Sarathchandran$^{*}$ \\
  TUDelft\\
  Netherlands \\
  \texttt{v.sarathchandran@tudelft.nl} \\
  \And
  Avijit Majhi$^{*}$ \\
  TUDelft\\
  Netherlands \\
  \texttt{a.majhi@tudelft.nl} \\
  \And
  Shao-Hsuan Hung \\
  TUDelft\\
  Netherlands \\
  \texttt{shaohung@tudelft.nl} \\
  \And
  Carlo Saccardi  \\
  TUDelft\\
  Netherlands \\
  \And
  Ruben Imhoff \\
  Deltares\\
  Netherlands \\
  \And
  Roberto Deidda  \\
  University of Cagliari\\
  Italy \\
  \And
  Remko Uijlenhoet  \\
  TUDelft\\
  Netherlands \\
  \And
  Justin Dauwels \\
  TUDelft\\
  Netherlands \\
}

\begin{document}

\maketitle

\begin{abstract}
Predicting precipitation maps is a highly complex spatiotemporal modeling task, critical for mitigating the impacts of extreme weather events. Short-term precipitation forecasting, or nowcasting, requires models that are not only accurate but also computationally efficient for real-time applications. Current methods, such as token-based autoregressive models, often suffer from flawed inductive biases and slow inference, while diffusion models can be computationally intensive. 
To address these limitations, we introduce BlockGPT, a generative autoregressive transformer using batched tokenization (Block) method that predicts full two-dimensional fields (frames) at each time step. Conceived as a model-agnostic paradigm for video prediction, BlockGPT factorizes space–time by using self-attention within each frame and causal attention across frames; in this work, we instantiate it for precipitation nowcasting.
We evaluate BlockGPT on two precipitation datasets, viz. KNMI (Netherlands) and SEVIR (U.S.), comparing it to state-of-the-art baselines including token-based (NowcastingGPT) and diffusion-based (DiffCast+Phydnet) models. The results show that BlockGPT achieves superior accuracy, event localization as measured by categorical metrics, and inference speeds up to $31\times$ faster than comparable baselines. Here there is the official implemention of BlockGPT:\url{https://github.com/Cmeo97/BlockGPT}.
\end{abstract}
\section{Introduction}
Climate change is increasing the frequency and intensity of extreme rainfall worldwide, disrupting infrastructure and posing risks to life and property \cite{Alfieri2017GlobalPO, martinkova2020overview, Klocek2021MSnowcastingOP, Czibula2021AutoNowPAA, MalkinOndik2022Nowcasting}. This amplifies the need for accurate, high-resolution short-term weather forecasting (nowcasting) \cite{cote2015seamless}. In operational early-warning chains \cite{imhoff2020spatial,Imhoff2023PystepsBlending}, short-range forecasts are typically produced by numerical weather prediction (NWP) systems; however, for minute-to-hour lead times, practical constraints—latency, update frequency, and effective resolution—can hinder the timing and localization of intense rainfall \cite{bauer2015quiet,Berenguer2012DiurnalCycleII,Pierce2012NowcastingChapter}. Crucially, NWP and radar-based nowcasting are complementary: NWPs provide large-scale dynamical context and longer lead times, whereas radar nowcasting leverages high-resolution observations for within-event, local decision-making. Rainfall nowcasting classically denotes statistical/heuristic extrapolation of real-time quantitative precipitation estimates (QPEs), exploiting radar’s fine spatiotemporal resolution (often $\sim$1~km/5~min) and direct initialization from the latest observations \cite{overeem2009derivation}. Methods include (i) field-based advection with stochastic evolution \cite{seed2003dynamic,bowler2006steps,Berenguer2011SBMcast,seed2013formulation,Sokol2017ProbNowcasting,ayzel2019optical}, (ii) object-oriented cell tracking \cite{Dixon1993TITAN,Han2009EnhancedTITAN}, (iii) analogue approaches \cite{Atencia2014LET,atencia2015comparison}, and (iv) machine learning \cite{shi2015convolutional,ravuri2021skilful,Luo2021PredRANNTS,Liu2022ASD}. Community efforts such as \texttt{pysteps} have consolidated and advanced these approaches in open source \cite{pulkkinen2019pysteps}. Recent work reframes radar nowcasting as a video prediction task, learning to propagate precipitation fields over minute-to-multi-hour horizons with low latency \cite{shi2015convolutional,ravuri2021skilful,prudden2020review}. In practice, data-driven nowcasts guide local decisions at short leads ($\approx$0–3~h), while NWP supplies the large-scale dynamics for longer horizons and basin- to synoptic-scale planning \cite{bauer2015quiet}. This perspective motivates modern generative sequence models for video prediction. State-of-the-art (SOTA) approaches employ VQ-VAEs \cite{van2017neural, meo2024alphatcvae}, transformers \cite{vaswani_attention_2017, meo_extreme_2024, Bi2023Nowcasting, Yin2024Precipitation}, and diffusion models \cite{gao2023prediff, yu_diffcast_2024} to improve accuracy and efficiency. Despite progress, long-term consistency, latency, and computational cost remain key challenges. To address these limitations, we introduce BlockGPT, a transformer that models spatiotemporal rainfall dynamics via frame-level autoregression. By predicting entire precipitation fields at each step, BlockGPT avoids the inductive biases and computational bottlenecks of token-level autoregression, yielding more coherent predictions and faster inference. Our contributions are: (1) a generative transformer that autoregressively predicts full precipitation fields, contrasting with prior token-level approaches; (2) a comprehensive evaluation on SEVIR \cite{veillette2020sevir} and KNMI \cite{knmi_dataset} showing state-of-the-art categorical skill and event localization, with inference up to 31$\times$ faster than SOTA baselines.

\section{Related Work}
Early deep learning efforts for nowcasting relied on recurrent neural networks (RNNs) \cite{rumelhart1986learning} architectures to model temporal sequences. Models like ConvLSTM \cite{shi2015convolutional} and ConvGRU \cite{shi2017deep} adapted RNNs designed for spatio-temporal data by replacing matrix multiplications with convolutional operations. This line of work was extended by models such as TrajGRU \cite{shi2017deep}, which improved motion tracking, and PhyDNet \cite{guen2020disentangling}, which embedded physical constraints by decomposing the latent space. DGMR \cite{ravuri2021skilful} employed a Generative Adversarial Network (GAN) \cite{goodfellow2014generative} with spatial and temporal discriminators to improve forecast quality. More recently, diffusion models \cite{ho2020denoising} have become prominent because of their stable training and high-fidelity generation. Models like PreDiff \cite{gao2023prediff} perform denoising in a latent space to generate future frames. However, it requires more than 30 days to be trained. In contrast, the proposed BlockGPT can be trained in 1.5 hours. DiffCast+Phydnet \cite{yu_diffcast_2024}, a key baseline in our work, introduced a residual diffusion approach where a base model predicts a coarse forecast and a diffusion model learns to predict the stochastic residual. Transformers offer an alternative to recurrent models that has been shown to be more stable, efficient, and scalable, leveraging self-attention to capture long-range dependencies \cite{vaswani_attention_2017, meo2025maskedgenerativepriorsimprove}. MAU \cite{chang2021mau}, for example, integrates motion cues through temporal aggregation, while Earthformer \cite{gao2022earthformer} applies cuboidal self-attention over radar volumes. Most closely related to our work is NowcastingGPT \cite{meo_extreme_2024}, which tokenizes radar precipitation fields using a VQ-VAE \cite{van2017neural} and autoregressively predicts them with a transformer decoder. However, the autoregression in NowcastingGPT operates at the token level rather than in time, creating an ill-posed learning problem that results in fragmented outputs and slow inference. To address these limitations, we propose BlockGPT, a generative transformer model that predicts entire precipitation fields at once in latent space, employing a block attention mask to enable bidirectional spatial attention within each precipitation field while maintaining temporal causality across precipitation fields. Motivated readers can also find a releted work section about video prediction architectures in appendix \ref{appendix:related_work_vp}.
\section{Methodology: BlockGPT Pipeline}
Given a sequence of $T_c$ context precipitation fields $X_{1:T_c}$, with $X_t \in \mathbb{R}^{H\times W}$, where ${H\times W}$ is the grid size of the rainfall fields, and the task is to predict the following $T$ future precipitation fields. The proposed BlockGPT pipeline, decomposes the prediction task into two stages: (1) compressing precipitation fields into a latent token space, and (2) autoregressively modelling temporal dynamics. 
\vspace{-0.65 cm}
\paragraph{Feature Exctraction}Each precipitation field $X_t$ is encoded using a VQ-GAN \cite{esser2021taming}, which downsamples and discretises it into a grid of latent tokens $\boldsymbol{\mathcal{T}}_t \in \mathbb{R}^{H' \times W'}$. A detailed description of the employed VQ-GAN and its training procedure can be found in appendix \ref{appendix:model}.
\vspace{-0.4 cm}
\paragraph{Dynamics Modeling}Prior work such as NowcastingGPT \cite{meo_extreme_2024} models temporal evolution by flattening each grid $\boldsymbol{\mathcal{T}}_t$ into a 1D token sequence $\mathbf{z}_t$, and concatenating across time $\mathbf{z}_t  = \left( z_t^{(1)}, z_t^{(2)}, \dots, z_t^{(H'W')} \right)$,
with the joint distribution factorised autoregressively as:
\vspace{-0.3 cm}
\begin{equation}    
p(\mathbf{z}) = \prod_{i=1}^{T \cdot H'W'} p\left(z^{(i)} \mid z^{(1)}, z^{(2)}, \dots, z^{(i-1)}\right).
\label{eq:factor1}
\end{equation}
However, such formulation implicitly assumes a sequential correlation of all $z^{(t)}$, imposing a flawed inductive bias. Indeed, the spatial tokens $z^{(t)}$ within a precipitation field are bidirectionally correlated and not naturally sequential. Treating them as a causal chain forces the model to predict inherently co-dependent tokens autoregressively, which is an ill-posed modelling assumption and leads to inefficient, suboptimal decoding. By contrast, we factorise the joint distribution as:
\begin{equation}  
p(\boldsymbol{\mathcal{T}}) = \prod_{t=1}^{T} p\left( \boldsymbol{\mathcal{T}}_t \mid \boldsymbol{\mathcal{T}}_1, \dots, \boldsymbol{\mathcal{T}}_{t-1} \right),
\label{eq:factor2}
\end{equation}
where each $\boldsymbol{\mathcal{T}}_t$ is holistically modelled, preserving the original 2D structure that contains bidirectionally correlated features. This distribution shift has several advantages, the first and most important is fixing the flawed inductive bias - features are not anymore assumed to be sequentially correlated. Secondly, the autoregression step is performed at the actual time scale, which allows the model to learn a meaningful time dependent dynamics. Finally, inference is $H'W'$ times faster by design, since we can now infer a complete $\boldsymbol{\mathcal{T}}_t$ with a single forward pass. 
It is important to note that, during training and inference, we use block attention masks - spatial tokens within a precipitation field are allowed to attend bidirectionally, while temporal attention is strictly causal. This design more naturally aligns with the spatiotemporal structure of precipitation: radar maps require full-field spatial modeling, whereas future precipitation fields should depend only on the past context. 
\section{Experiments}

We evaluate all models on the task of nowcasting, where the goal is to predict the next 6 radar precipitation fields given 3 context precipitation fields. Each precipitation field represents 30 minutes of precipitation, resulting in a forecast horizon of 3 hours. Experiments are conducted on two real-world radar datasets: the Dutch KNMI dataset \cite{knmi_dataset} and the SEVIR dataset \cite{veillette2020sevir} from the United States. Appendix \ref{appendix:datasets} presents analyses of the considered datasets, providing a detailed overview of the datasates statistics. 
We benchmark BlockGPT against NowcastingGPT \cite{meo_extreme_2024} and DiffCast+Phydnet \cite{yu_diffcast_2024}. To the best of our knowledge, the former is the current state-of-the-art discrete token-based autoregressive model for precipitation nowcasting in the KNMI dataset, while the latter exemplifies the residual diffusion paradigm \cite{yu_diffcast_2024} and is the state-of-the-art in the SEVIR dataset. We report quantitative results in terms of four key metrics: Mean Squared Error (MSE), Pearson Correlation Coefficient (PCC), Critical Success Index (CSI), and False Alarm Rate (FAR). Details about the experiments can be found in Appendix \ref{appendix:setup}.
\begin{figure}[h]
    \centering
    \includegraphics[width=0.8\linewidth]{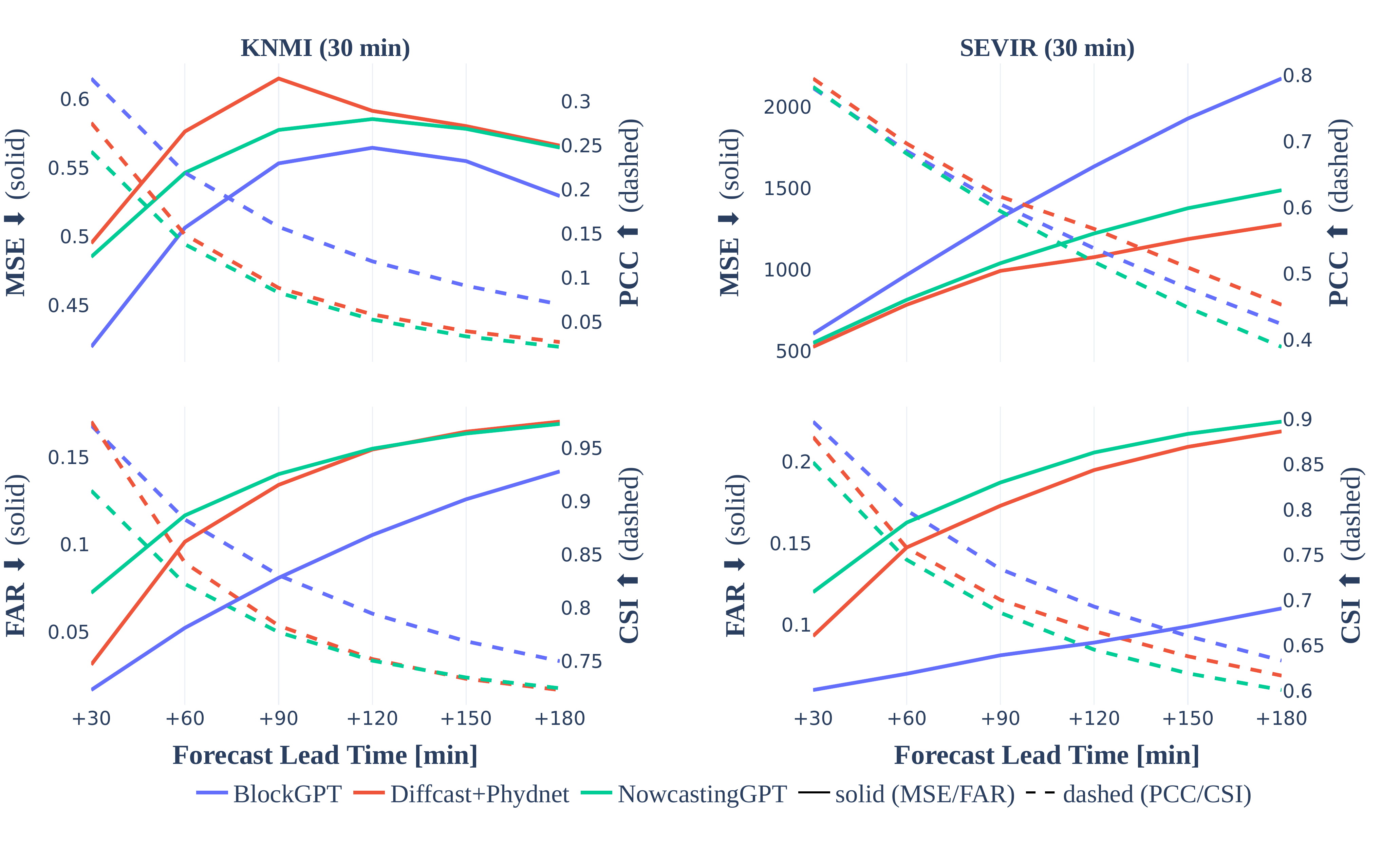}%
    \caption{MSE, PCC, CSI, and FAR of BlockGPT and related baselines, on KNMI and SEVIR datasets. Results are averaged across 3 seeds. \vspace{-0.5 cm}}
    \label{fig:results_summary}
\end{figure}

\subsection{Results}
In this section, we design empirical experiments to understand the performance of BlockGPT and its
potential limitations by exploring the following questions: (1) How does BlockGPT's frame-level autoregressive approach compare to token-level autoregression (NowcastingGPT) and diffusion-based models (DiffCast+Phydnet) in terms of prediction accuracy and computational efficiency? (2) What differences are there between autoregressive and diffusion-based generative behaviors?
\vspace{-0.35 cm}
\paragraph{Block Autoregressive models are better nowcasters than diffusion models.}
Figure \ref{fig:results_summary} presents a comparative performance analysis of three models: Diffcast+Phydnet, BlockGPT, and NowcastingGPT on the KNMI and SEVIR datasets for forecast lead times up to 180 minutes. A consistent performance hierarchy is evident across both datasets, all forecast lead times and all metrics, with BlockGPT outperforming Diffcast+Phydnet and NowcastingGPT, with the only exception for MSE and PCC evaluation on SEVIR where DiffCast+Phydnet performs the best (Diffcast + Phydnet: $\sim$ 1250 $\rightarrow$ NowcastingGPT: $\sim$ 1450 $\rightarrow$ BlockGPT: $\sim$ 2100). However, BlockGPT shows superior event-level detection performance as reflected in higher CSI and lower FAR values. Qualitative results can be found in Appendix \ref{appendix:qualitativ-studies} and the percentile-wise continuous metrics in Appendix \ref{appendix:continuous-score-percentile-levels}. As illustrated by the qualitative case studies in Appendix \ref{appendix:qualitativ-studies}, BlockGPT preserves storm morphology and displacement more faithfully than the baselines across KNMI and SEVIR events (see Fig. \ref{fig:knmi_viz_exp1}, \ref{fig:knmi_viz_exp2}, \ref{fig:sevir_viz_exp1}, \ref{fig:sevir_viz_exp2}). On SEVIR, BlockGPT occasionally overestimates high intensity storm cells at longer lead times—consistent with the continuous scores reported in App. \ref{appendix:continuous-score-percentile-levels} (see Fig. \ref{fig:metrics_all_levels_sevir30}), whereas on KNMI, errors decrease toward over the leadtimes and PCC remains systematically higher (see Fig. \ref{fig:metrics_all_levels_knmi30}).
\vspace{-0.45 cm}
\paragraph{BlockGPT is more robust than diffusion pipelines.} 
To further evaluate model performance at various rainfall intensities, we also report the Area Under the ROC Curve (AUC) over time for different precipitation thresholds on the KNMI dataset in Fig. \ref{fig:knmi_roc}. BlockGPT consistently outperforms both baselines at all thresholds and time steps, demonstrating robustness in detecting precipitation events of varying severity. Consistent trends are observed at the catchment scale in \ref{appendix:catchment-analysis}, where AUC-ROC computed for 1, 2, and 8\,mm\,h$^{-1}$ across +30 to +180\,min lead times confirms that BlockGPT maintains the highest detection skill across thresholds (\autoref{fig:knmi_roc}).
\section{Conclusion}
In this work, we introduced BlockGPT, a frame-level autoregressive transformer designed for precipitation nowcasting. By shifting the generative process from a token-by-token sequence to predicting entire precipitation fields autoregressively, BlockGPT overcomes the flawed inductive biases and computational bottlenecks of prior token-level models, resulting in a 31x faster inference speed than its counterparts, as showed in Appendix \ref{appendix:computational_requirements} and \ref{app:params}. This approach enables the use of bidirectional attention to capture complex spatial patterns within each frame while strictly maintaining temporal causality across frames.
Our comprehensive evaluation on the KNMI and SEVIR datasets demonstrates that BlockGPT consistently outperforms state-of-the-art models on key categorical metrics, including the CSI and Area Under the ROC Curve. This indicates a stronger ability to accurately localize and predict precipitation events, particularly those exceeding critical intensity thresholds. 
Future work could focus on enhancing fine-grained prediction accuracy, potentially by integrating BlockGPT as a powerful backbone within a residual diffusion framework. Further research could also explore the incorporation of physical constraints and the development of robust uncertainty quantification methods to improve prediction reliability for critical decision-making.

\section*{Acknowledgments}

This work is part of the project Delft AI4WF: Delft Artificial Intelligence for Weather Forecast with file number 2024.023 of the research programme Computing time of national computer systems, which is (partly) financed by the NWO under the grant \url{https://doi.org/10.61686/VYGRS56933}.

\newpage
\bibliography{bibliography}

\appendix
\newpage
\section{Dataset Analysis}
\label{appendix:datasets}
\subsection{KNMI Dataset}

The KNMI dataset contains radar-based precipitation estimates collected by two weather radars located in the Netherlands~\cite{knmi_dataset}. The data has a spatial resolution of 1~km$^{2}$ and a temporal resolution of 5 minutes, covering the entire land area of the Netherlands. It spans the period from 2008 to 2018, during which the radar infrastructure underwent a significant upgrade.

The raw measurements recorded by these radars are in the form of \textit{radar reflectivity}, which quantifies the amount of transmitted microwave energy reflected back after encountering precipitation particles. Reflectivity values are converted to precipitation rates using a standard Z–R relationship, given by:
\[
Z_h = 200 R^{1.6},
\]
where \( Z_h \) is the horizontal radar reflectivity factor and \( R \) is the precipitation rate in mm/hr~\cite{marshall1955advances}.

The dataset is highly imbalanced towards low/no precipitation events. To address this, we retain only those events with average precipitation above the 50th percentile for all subsequent experiments. This subset includes data spanning from 2008 to 2018 and provides a more informative, balanced foundation for model training and evaluation.

\begin{figure}[H]
    \centering
    \includegraphics[width=\linewidth]{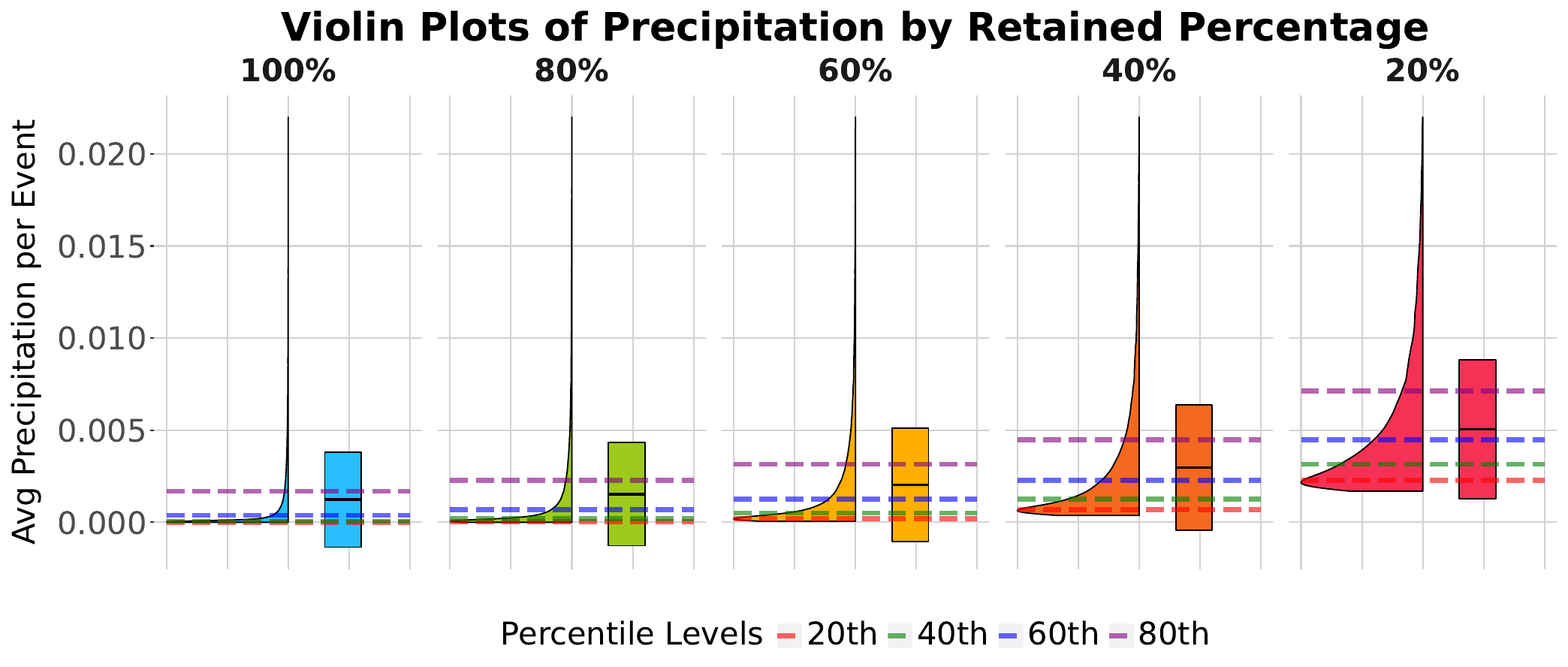}
    \caption{Violin plots of event average precipitation in the KNMI dataset.}
    \label{fig:violin_knmi_app}
\end{figure}
\subsection{SEVIR Dataset}

The SEVIR dataset, introduced in~\cite{veillette2020sevir}, is a machine-learning-ready resource that aggregates multiple remote sensing modalities, including satellite and radar data. It consists of 4-hour weather events covering 384~km~$\times$~384~km regions across the continental United States, sampled every 5 minutes. The spatial resolution is 1~km$^{2}$ for most modalities, including Vertically Integrated Liquid (VIL).

SEVIR provides five sensing modalities: three channels from the GOES-16 (Geostationary Operational Environmental Satellite) system~\cite{schmit2017closer}, VIL measurements, and data from the Geostationary Lightning Mapper (GLM)~\cite{goodman2013goes}. VIL, which is derived from radar reflectivity, estimates the total liquid water content in a vertical column of the atmosphere and is commonly used to assess intense precipitation events such as thunderstorms and hail~\cite{nws_vil_density}. For this study, we focus on the VIL modality.

\begin{figure}[H]
    \centering
    \includegraphics[width=\linewidth]{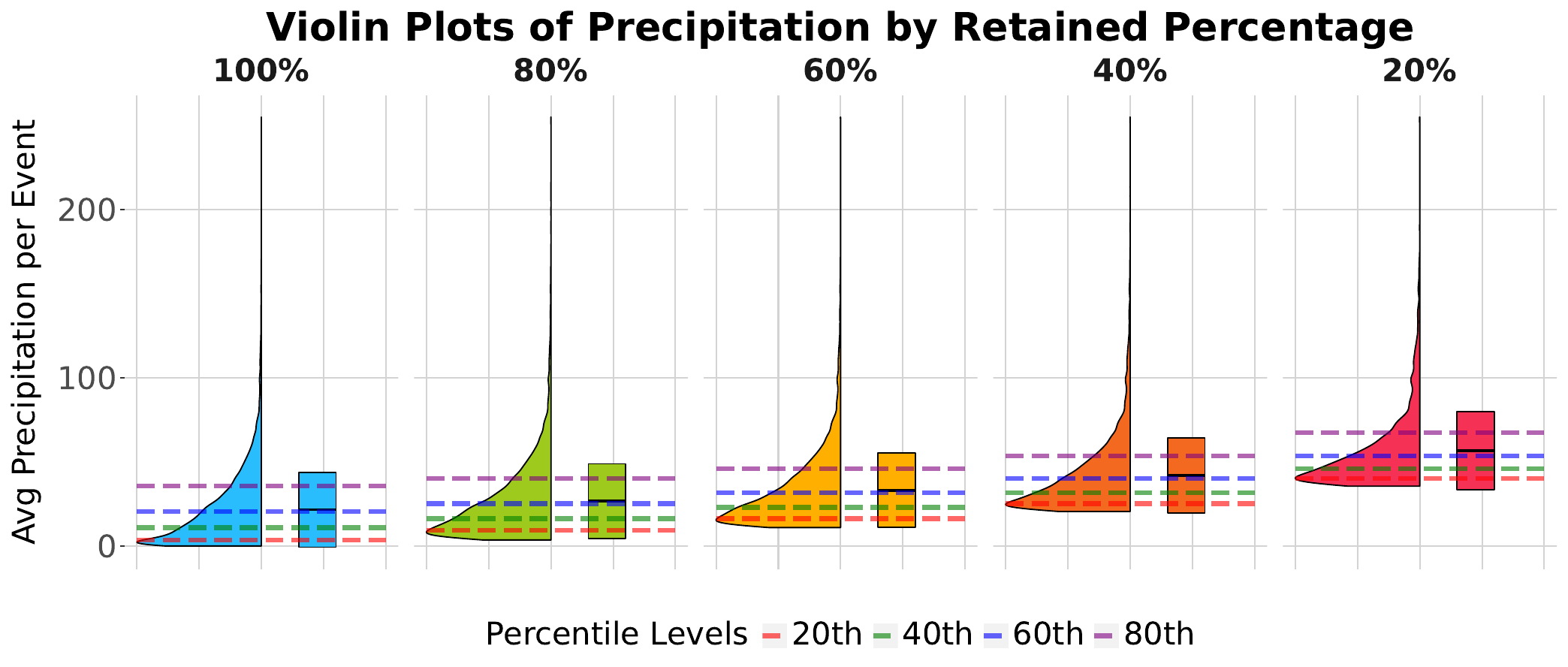
}
    \caption{Violin plots of event average precipitation in the SEVIR dataset.}
    \label{fig:violin_sevir_app}
\end{figure}

\include{extended_results}

\begin{figure}[h]
    \centering
    \includegraphics[width=1\linewidth]{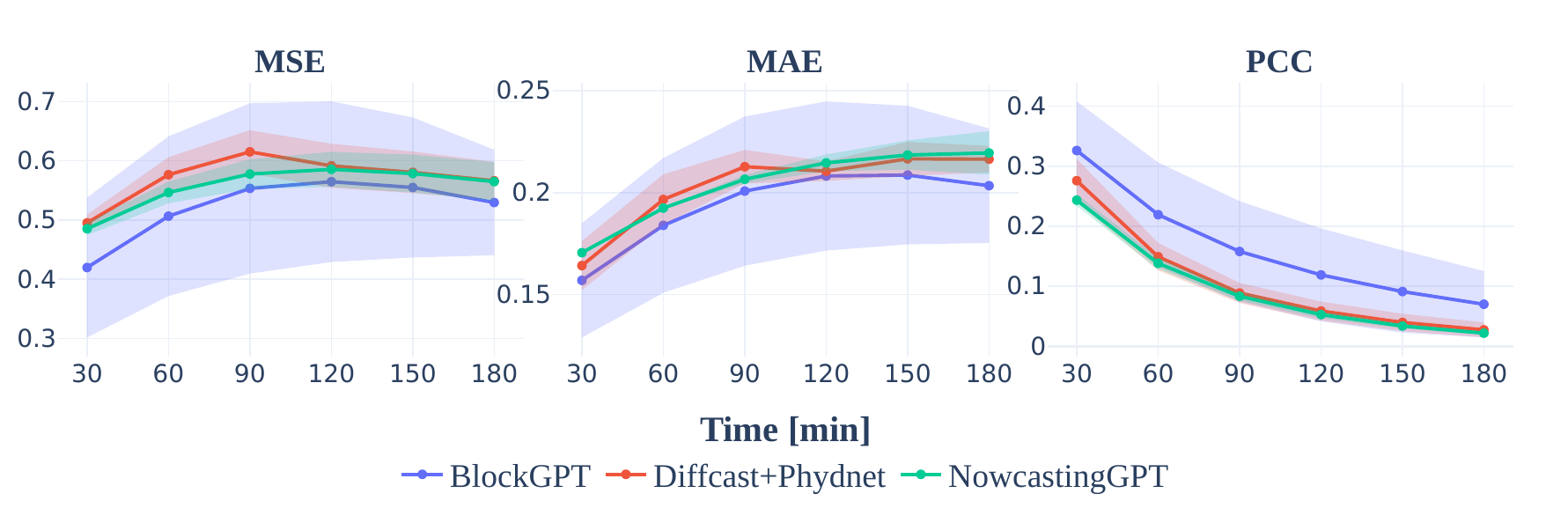}
    \caption{KNMI: Continuous metrics (MSE, MAE, PCC) by percentile bin (0--20, 20--40, 40--60, 60--80, 80--95th). BlockGPT exhibits superior PCC across bins; MSE/MAE reduce notably at higher bins. Aggregated scores favour BlockGPT overall.}
    \label{fig:metrics_all_levels_knmi30}
\end{figure}
\begin{figure}[h]
    \centering
    \makebox[\linewidth][l]{%
        \hspace*{-0.5cm}%
        \includegraphics[width=1\linewidth]{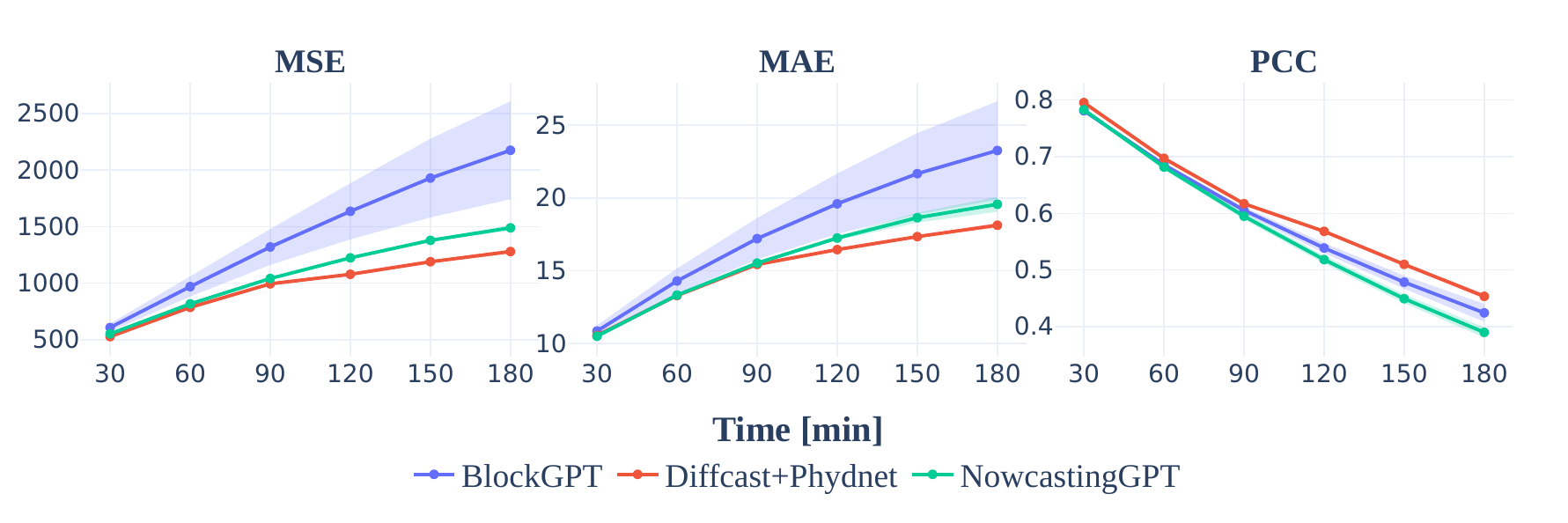}%
    }
    \caption{SEVIR: Continuous metrics (MSE, MAE, PCC) by percentile bin (0--20, 20--40, 40--60, 60--80, 80--95th). BlockGPT shows elevated MSE/MAE due to overestimation of high-intensity cores at long lead times, while maintaining favourable PCC relative to DiffCast+Phydnet.}
    \label{fig:metrics_all_levels_sevir30}
\end{figure}

\section{Model Architecture Details}
\label{appendix:model}

\subsection{VQ-GAN Training Details}

The first stage of the BlockGPT pipeline compresses high-dimensional precipitation maps into a compact, discrete latent representation. For this, we employ a Vector Quantized-Generative Adversarial Network (VQ-GAN) \cite{esser2021taming}, which leverages an encoder-decoder framework with a discrete codebook $\mathcal{Z}$, and a discriminator $\mathcal{D}$, that discriminates between reconstructed and ground truth images. The encoder maps an input precipitation map $x$ into a lower-resolution grid of feature vectors, preserving essential spatial information by reducing spatial dimensions while increasing feature channels. Specifically, the encoder consists of $5$ downsampling layers, each containing $2$ ResNet blocks, which progressively reduce the spatial resolution from $128 \times 128$ down to $8 \times 8$. The final stage of the encoder includes an attention block to better capture global feature relationships before quantization.

Each feature vector $\hat{z}= E(x)$ is then mapped to its closest entry in the learned codebook $\mathcal{Z}$ via an element-wise quantization step $q(\cdot)$:
$$
z_{\mathbf{q}}=\mathbf{q}(\hat{z}):=\left(\underset{z_k \in \mathcal{Z}}{\arg \min }\left\|\hat{z}-z_k\right\|\right).
$$
This process yields a grid of discrete latent tokens $z_q$ for each input frame. The decoder, which mirrors the encoder's architecture, then reconstructs the precipitation map $\hat{x} = \text{Dec}(z_q)$ from these quantized tokens. The VQ-GAN is trained by optimizing a combination of reconstruction, commitment, and perceptual losses to ensure both high-fidelity reconstruction and a well-structured latent space:
\begin{equation}
\begin{aligned}
\mathcal{L}_{\text{VQ-VAE}} = \|x-\hat{x}\|_2^2 + \beta \left\| \text{sg}[E(x)]-z_{\mathbf{q}} \right\|_2^2 + \left\| \text{sg}[z_{\mathbf{q}}]-E(x) \right\|_2^2 + \mathcal{L}_{\text{perceptual}}(x, \hat x),
\end{aligned}
\label{eq:VQVAE}
\end{equation}
where $\text{sg}[\cdot]$ denotes the stop-gradient operator, and the commitment loss (the third term) is weighted by a hyperparameter $\beta$.
To ensure the generation of realistic maps, an adversarial loss from the discriminator $\mathcal{D}$ is added. Therefore, the GAN Loss ($\mathcal{L}_{\text{GAN}}$) loss is given by:
\begin{equation}
\mathcal{L}_{\text{GAN}} = \mathcal{L}_{\text{VQ-VAE}} + \lambda\mathbb{E}_{x\sim p(x)}[\log \mathcal{D}(x)+\log(1-\mathcal{D}(\hat{x}))]
\label{eq:VQGAN}
\end{equation}
where the term $\lambda$ is an adaptive weight calculated from the gradients of the perceptual and GAN losses to balance their contributions during training.

\section{Extended Related Work}
Precipitation nowcasting, a sub-field of spatio-temporal forecasting, presents unique challenges due to the chaotic and stochastic evolution of weather systems. While traditional methods based on physical principles, like the advection-based PySTEPS \cite{pulkkinen2019pysteps}, are well-established, they often struggle to model complex, non-linear dynamics \cite{ravuri2021skilful}. In contrast, deep learning (DL) models have demonstrated a remarkable ability to learn these patterns directly from vast amounts of radar data \cite{shi2015convolutional, ravuri2021skilful}.
The shift of paradigm that led DL models to succeed was casting precipitation nowcasting as a video prediction problem \cite{Bi2023Nowcasting, Bai2022RainformerFE, Luo2021PredRANNTS}, where given an input spatio-temporal sequence of $N$ frames $\boldsymbol{x}_{\text{in}} \in \mathbb{R}^{N \times H \times W \times C}$, $H, W$ denote the spatial resolution and $C$ represents the image channels or the different type of measurements (e.g., radar, heat maps, etc), the goal is to predict the next $M$ frames $\boldsymbol{x}_{\text{out}} \in \mathbb{R}^{M \times H \times W \times C}$. In this section we present the literature related to precipitation nowcasting models, the main related field of this paper. 
\subsection{Video Prediction Architectures}
\label{appendix:related_work_vp}

A prominent architectural pattern in modern generative video modeling involves a three-stage process: (1) a compression stage that encodes high-dimensional frames into a discrete latent space, (2) a generation stage that models the dynamics of these latent representations, and (3) a diffusion step that models the residuals that were not captured by the video prediction backbone. 

The compression is typically handled by a Vector Quantized-Variational Autoencoder (VQ-VAE) \cite{van2017neural}, which learns a codebook of visual tokens. The generation is then performed by a powerful sequence model, often an autoregressive Transformer \cite{vaswani_attention_2017}, which learns to predict the next token in a sequence. This approach was popularized for general video generation by VideoGPT \cite{yan2021videogpt} and adapted for precipitation nowcasting by NowcastingGPT \cite{meo_extreme_2024}. These models typically flatten the 2D grid of spatial tokens into a 1D sequence and predict them one-by-one.

However, this token-level autoregression imposes a flawed inductive bias by assuming a causal, sequential relationship between tokens that are spatially correlated. This creates an ill-posed learning problem that can result in spatially fragmented outputs and suffers from slow inference speeds due to its sequential nature \cite{tian2024visual, luo2024open}.

To address these limitations, recent works have shifted towards generating tokens in larger chunks or in parallel. Some methods have explored non-autoregressive generation using masking strategies \cite{chang2022maskgit}. Our work is most closely related to the emerging paradigm of block-level autoregression, where an entire block of tokens—or in our case, an entire frame—is predicted at each time step \cite{li2024autoregressive, yu2023language}. This approach, explored in models like Next Block Prediction \cite{ren2025next}, allows for bidirectional self-attention within a frame to capture spatial structures effectively, while maintaining a causal autoregressive structure across time to model temporal evolution. By adopting this frame-level prediction strategy, BlockGPT aims to overcome the efficiency and coherence issues of prior token-based nowcasting models.

\section{Experimental Setup Details}
\label{appendix:setup}

\subsection{Training Configuration}
All models are trained with Adam (learning rate $1\times 10^{-4}$) using a 10{,}000-step warmup. Training runs for 500{,}000 steps with batch size 8. Evaluation is performed on a held-out test set unused during training/validation.

\subsection{Evaluation Metrics}
To assess the quality of predicted precipitation sequences, we employ continuous (value-based) and categorical (event-based) metrics. Let an event be a sequence of $T$ frames
\[
\mathcal{X}=\{\mathbf{X}_1,\dots,\mathbf{X}_{T_c},\mathbf{X}_{T_c+1},\dots,\mathbf{X}_T\}
=\{\mathcal{X}_{\text{context}},\mathcal{X}_{\text{target}}\},
\]
where $\mathbf{X}_t\in\mathbb{R}^{H\times W}$ is the radar field at time $t$. Given predictions $\hat{\mathcal{X}}_{\text{target}}=\{\hat{\mathbf{X}}_{T_c+1},\dots,\hat{\mathbf{X}}_T\}$ and targets $\mathcal{X}_{\text{target}}=\{\mathbf{X}_{T_c+1},\dots,\mathbf{X}_T\}$, we define:

\subsubsection*{Continuous Metrics}
These quantify \cite{jolliffe2012forecast} amplitude accuracy over the full spatiotemporal domain.

\textbf{Mean Squared Error (MSE).} Penalizes large deviations quadratically; sensitive to outliers and thus captures severe intensity errors.
\begin{equation}
\text{MSE}=\frac{1}{(T-T_c)HW}\sum_{t=T_c+1}^{T}\sum_{i=1}^{H}\sum_{j=1}^{W}\bigl(\hat{\mathbf{X}}_t[i,j]-\mathbf{X}_t[i,j]\bigr)^2.
\end{equation}

\textbf{Mean Absolute Error (MAE).} Measures median-like deviation; robust to outliers and interpretable in physical units.
\begin{equation}
\text{MAE}=\frac{1}{(T-T_c)HW}\sum_{t=T_c+1}^{T}\sum_{i=1}^{H}\sum_{j=1}^{W}\bigl|\hat{\mathbf{X}}_t[i,j]-\mathbf{X}_t[i,j]\bigr|.
\end{equation}

\textbf{Pearson Correlation Coefficient (PCC).} Assesses linear association and phase coherence independent of bias and scale.
\begin{equation}
\text{PCC}=\frac{\sum_{t,i,j}\bigl(\hat{\mathbf{X}}_t[i,j]-\bar{\hat{\mathbf{X}}}\bigr)\bigl(\mathbf{X}_t[i,j]-\bar{\mathbf{X}}\bigr)}%
{\sqrt{\sum_{t,i,j}\bigl(\hat{\mathbf{X}}_t[i,j]-\bar{\hat{\mathbf{X}}}\bigr)^2}\;
 \sqrt{\sum_{t,i,j}\bigl(\mathbf{X}_t[i,j]-\bar{\mathbf{X}}\bigr)^2}},
\end{equation}
where $\bar{\hat{\mathbf{X}}}$ and $\bar{\mathbf{X}}$ are means over all target pixels and times.

\subsubsection*{Categorical Metrics}
These evaluate \cite{jolliffe2012forecast}event detection (e.g., exceedance of a threshold $\tau$). We binarize frames as
\begin{align}
\hat{\mathbf{Y}}_t[i,j]&=\mathds{1}\bigl[\hat{\mathbf{X}}_t[i,j]\ge\tau\bigr], \label{eq:binarize_hat}\\
\mathbf{Y}_t[i,j]&=\mathds{1}\bigl[\mathbf{X}_t[i,j]\ge\tau\bigr]. \label{eq:binarize_true}
\end{align}
Over all $(t,i,j)$ in the target period, the contingency counts are
\begin{align}
\mathrm{TP}&=\sum_{t,i,j}\mathds{1}\bigl[\hat{\mathbf{Y}}_t[i,j]=1,\ \mathbf{Y}_t[i,j]=1\bigr], \label{eq:tp}\\
\mathrm{FP}&=\sum_{t,i,j}\mathds{1}\bigl[\hat{\mathbf{Y}}_t[i,j]=1,\ \mathbf{Y}_t[i,j]=0\bigr], \label{eq:fp}\\
\mathrm{FN}&=\sum_{t,i,j}\mathds{1}\bigl[\hat{\mathbf{Y}}_t[i,j]=0,\ \mathbf{Y}_t[i,j]=1\bigr], \label{eq:fn}\\
\mathrm{TN}&=\sum_{t,i,j}\mathds{1}\bigl[\hat{\mathbf{Y}}_t[i,j]=0,\ \mathbf{Y}_t[i,j]=0\bigr]. \label{eq:tn}
\end{align}

\begin{table}[h]
\centering
\small
\begin{tabular}{c|cc}
 & \textbf{Observed 1 (event)} & \textbf{Observed 0 (no event)} \\\hline
\textbf{Forecast 1 (event)} & TP (hit) & FP (false alarm) \\
\textbf{Forecast 0 (no event)} & FN (miss) & TN (correct rejection) \\
\end{tabular}
\caption{Contingency table for threshold-exceedance events.}
\end{table}

\textbf{Critical Success Index (CSI).} Fraction of correctly predicted events among all observed or forecast events; penalizes misses and false alarms.
\begin{equation}
\text{CSI}=\frac{\mathrm{TP}}{\mathrm{TP}+\mathrm{FP}+\mathrm{FN}}.
\end{equation}

\textbf{False Alarm Ratio (FAR).} Proportion of forecast events that did not occur; lower is better.
\begin{equation}
\text{FAR}=\frac{\mathrm{FP}}{\mathrm{TP}+\mathrm{FP}}.
\end{equation}

\noindent\textbf{Receiver Operating Characteristic (ROC) and AUC.} The ROC curve \cite{jolliffe2012forecast,pulkkinen2019pysteps} measures discrimination skill for threshold-exceedance events by varying a decision threshold $\gamma$ on forecast probabilities or continuous scores derived from the predicted frames and comparing to the observed binary targets in \eqref{eq:binarize_true}. For each $\gamma$, compute the hit rate (probability of detection, POD) and the false-alarm rate (probability of false detection, POFD/FPR) from the contingency counts in \eqref{eq:tp}–\eqref{eq:tn}:
\begin{align}
\text{POD}(\gamma)&=\frac{\mathrm{TP}(\gamma)}{\mathrm{TP}(\gamma)+\mathrm{FN}(\gamma)}, \label{eq:pod}\\
\text{POFD}(\gamma)&=\frac{\mathrm{FP}(\gamma)}{\mathrm{FP}(\gamma)+\mathrm{TN}(\gamma)}. \label{eq:pofd}
\end{align}
Plotting $\text{POD}(\gamma)$ against $\text{POFD}(\gamma)$ yields the ROC curve; better discrimination pushes the curve toward the upper-left corner $(\text{POFD}=0,\text{POD}=1)$, indicating that predicted exceedances align with observed exceedances while rarely triggering on non-events (i.e., high hits, few false alarms). The area under the curve summarizes potential skill,
\begin{equation}
\text{AUC}=\int_{0}^{1}\text{POD}\bigl(\text{POFD}\bigr)\,d\,\text{POFD}, \label{eq:auc}
\end{equation}
with $\text{AUC}=0.5$ for no-skill and $\text{AUC}=1$ for perfect discrimination; note that $\text{POFD}$ in \eqref{eq:pofd} is not the same as FAR.

\subsection{Statistical Significance}
All scores are aggregated over 3 random seeds; where shown, shaded bands denote $\pm1$ standard deviation to convey sampling uncertainty.

\section{Model Parameters and Training Time}
\label{app:params}

In this section, we summarize the training configurations and compute profiles of the three models compared in this work. A key goal in our design of BlockGPT was to match or outperform the benchmark models in terms of training time, while scaling model capacity up to the point of overfitting. The table below presents a comparison of parameter counts and training durations across all models.

\begin{table}[h]
\centering
\begin{tabular}{lccc}
\toprule
Model & Parameters & Training Time & Hardware / Epochs \\
\midrule
DiffCast+Phydnet & 49.35M & $\sim$15 hours & 2 $\times$ A100 GPUs / 20 epochs \\
NowcastingGPT & 150M & $\sim$6 hours & 2 $\times$ A100 GPUs / 20 epochs \\
BlockGPT (Ours) & 103.37M & $\sim$6 hours & 2 $\times$ A100 GPUs / 20 epochs \\
\bottomrule
\end{tabular}
\caption{Training time and parameter comparison across all models.}
\label{tab:training_comparison}
\end{table}

Our model, BlockGPT, was designed under the constraint of maintaining a training budget that is no greater than that of our benchmarks. Within this constraint, we maximize model capacity by increasing the number of parameters up to the point of overfitting or until the training time matches that of the benchmarks. This approach ensures a fair and efficient comparison while allowing us to explore the benefits of larger model capacity within realistic computational limits. We retain the original model configurations of DiffCast+Phydnet and NowcastingGPT. For the latter, we retain the same model parameters as those in the checkpoints in the github repository of \cite{meo_extreme_2024}.  The embedding dimension however, was originally only 128. We therefore retrain with the same embedding dimension as ours, for fair comparison.

BlockGPT's model configuration is as follows: 

\begin{table}[H]
\centering
\caption{Essential architecture specifications for VQGAN and BlockGPT Transformer.}
\begin{tabular}{lll}
\toprule
Component & Parameter & Value \\
\midrule
VQGAN & Codebook Size & 1024 \\
      & Latent Channels & 128 \\
      & Attention Resolution & [8] \\
      & Dropout & 0.2 \\
      & Tokens per Frame & 64 \\
\midrule
BlockGPT (Transformer) & Number of Layers & 8 \\
                       & Number of Heads & 8 \\
                       & Embedding Size & 1024 \\
                       & Token Block Size & 576 \\
                       & Vocabulary Size & 1024 \\
\bottomrule
\end{tabular}
\label{tab:architecture_summary}
\end{table}

\section{Implementation Details}

\subsection{Code Availability}

The implementation of BlockGPT and all experimental code will be made publicly available upon publication. The codebase includes training scripts, evaluation metrics, and pre-trained models for reproducibility.

\subsection{Computational Requirements}
\label{appendix:computational_requirements}
Training BlockGPT requires approximately 8 GPUs with 32GB memory each for 500,000 steps. Inference can be performed on a single GPU, making it suitable for real-time applications.

A key advantage of BlockGPT is its computational efficiency. As shown in Table \ref{tab:inference_times}, BlockGPT is significantly faster than both benchmarks. On the 30-minute task, it is 27× faster than NowcastingGPT and 31× faster than DiffCast+Phydnet. On the 5-minute task, it is 31× faster than NowcastingGPT and 10× faster than DiffCast+Phydnet. These results indicate that frame-level autoregression not only improves performance but also greatly enhances computational efficiency.

\begin{table}[H]
\centering
\begin{tabular}{lc}
\toprule
Model & Inference Time (s) \\
\midrule
NowcastingGPT & 7.09 \\
DiffCast+Phydnet & 8.17 \\
BlockGPT & 0.26 \\
\bottomrule
\end{tabular}
\caption{Inference time (in seconds) per batch for each model.}
\label{tab:inference_times}
\end{table}

\subsection{Hyperparameter Tuning}

We conducted extensive hyperparameter tuning for all models to ensure fair comparison. The final hyperparameters were selected based on validation performance on a held-out validation set.

\end{document}

%% file: extended_results.tex
\section{Extended Results}
\label{appendix:extended-results}
\subsection{Qualitative Case Studies}
\label{appendix:qualitativ-studies}
We compare two representative events per dataset, each showing two input frames ($t=-60$ and $t=0$ min) and four forecast frames ($t=+30,+60,+120,+180$ min) from BlockGPT (ours), NowcastingGPT (token-level autoregression), and DiffCast+Phydnet (diffusion-based), against the ground truth.

\begin{figure}[h]
    \centering
    \makebox[\linewidth][l]{%
        \hspace*{-0.5cm}%
        \includegraphics[width=1.1\linewidth]{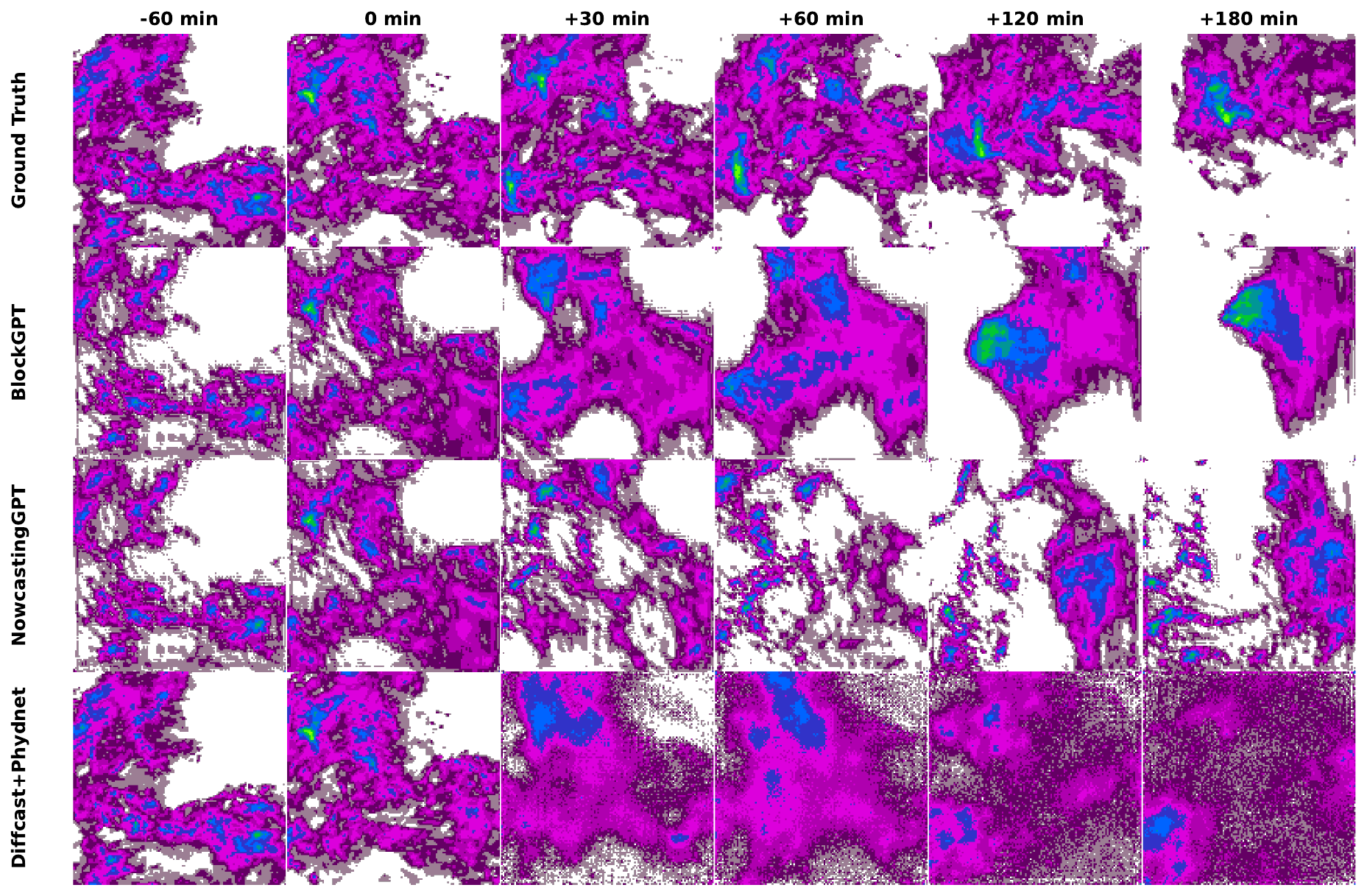}%
    }
    \caption{KNMI Event~1. Two input frames ($-60$, $0$ min) and four forecasts ($+30$, $+60$, $+120$, $+180$ min). BlockGPT preserves the rainband morphology and advection but modestly overestimates the core intensity at long lead times; baselines miss the shape and location.}
    \label{fig:knmi_viz_exp1}
\end{figure}

\begin{figure}[h]
    \centering
    \makebox[\linewidth][l]{%
        \hspace*{-0.5cm}%
        \includegraphics[width=1.1\linewidth]{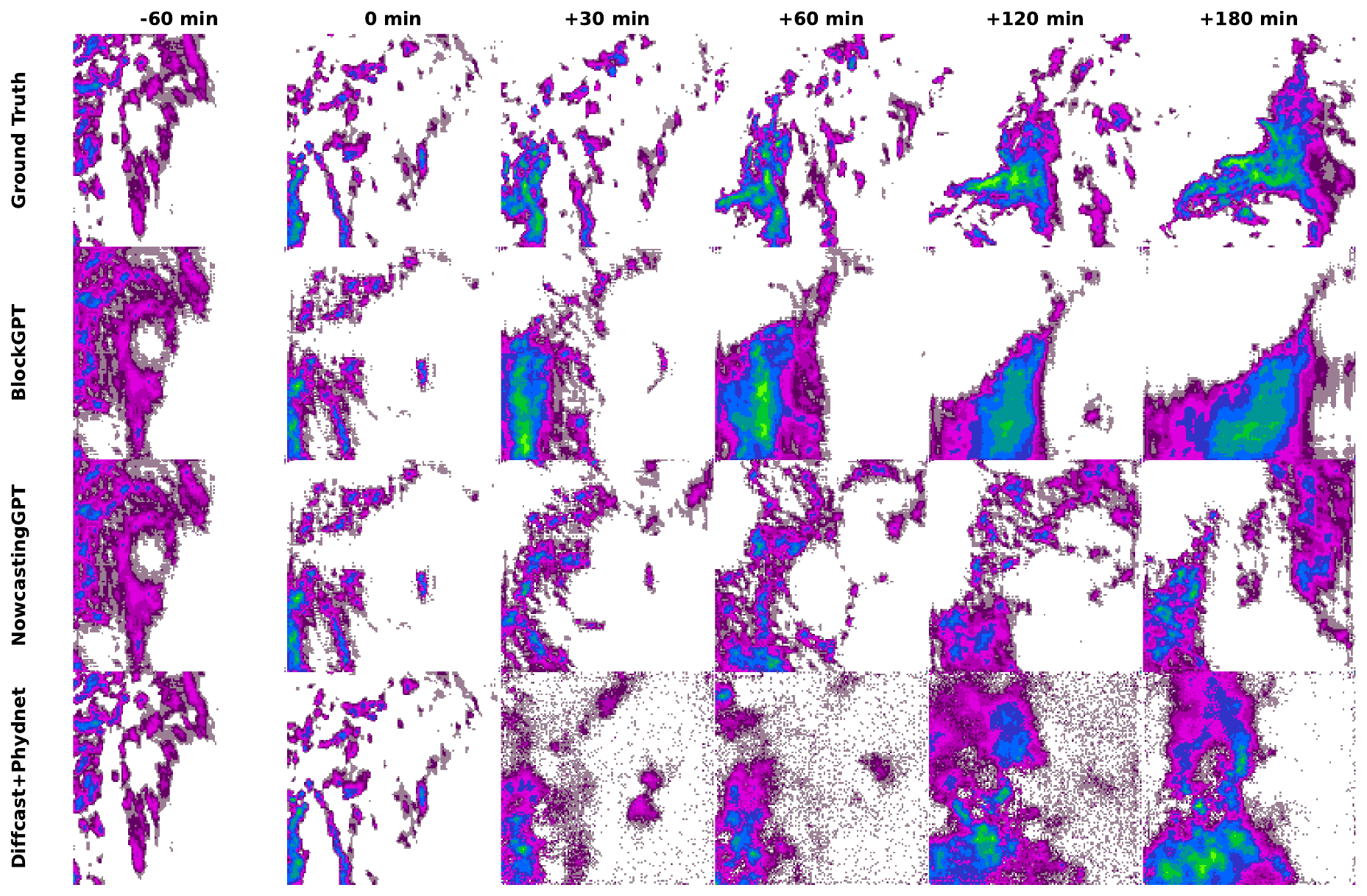}%
    }
    \caption{KNMI Event~2. BlockGPT follows the rapid structural changes and localisation of intense cells across lead times; baselines underperform, particularly for growth and displacement.}
    \label{fig:knmi_viz_exp2}
\end{figure}

\paragraph{KNMI.}
In \autoref{fig:knmi_viz_exp1}, BlockGPT captures the elongated rainband and its left-to-right advection, preserving embedded convective cores and their growth. While peak intensity is slightly overestimated at longer horizons, the morphology and displacement remain accurate. In contrast, NowcastingGPT and DiffCast+Phydnet miss the band structure and misplace intense cells; the diffusion pipeline also exhibits blobby artefacts and fails to recover the linear organisation. A similar pattern holds for the more challenging convective case in \autoref{fig:knmi_viz_exp2}, where rapid growth and relocation of high-intensity cells occur: BlockGPT reconstructs the evolving structure and localisation, whereas baselines struggle with both evolution and displacement.

\begin{figure}[h]
    \centering
    \makebox[\linewidth][l]{%
        \hspace*{-0.5cm}%
        \includegraphics[width=1.1\linewidth]{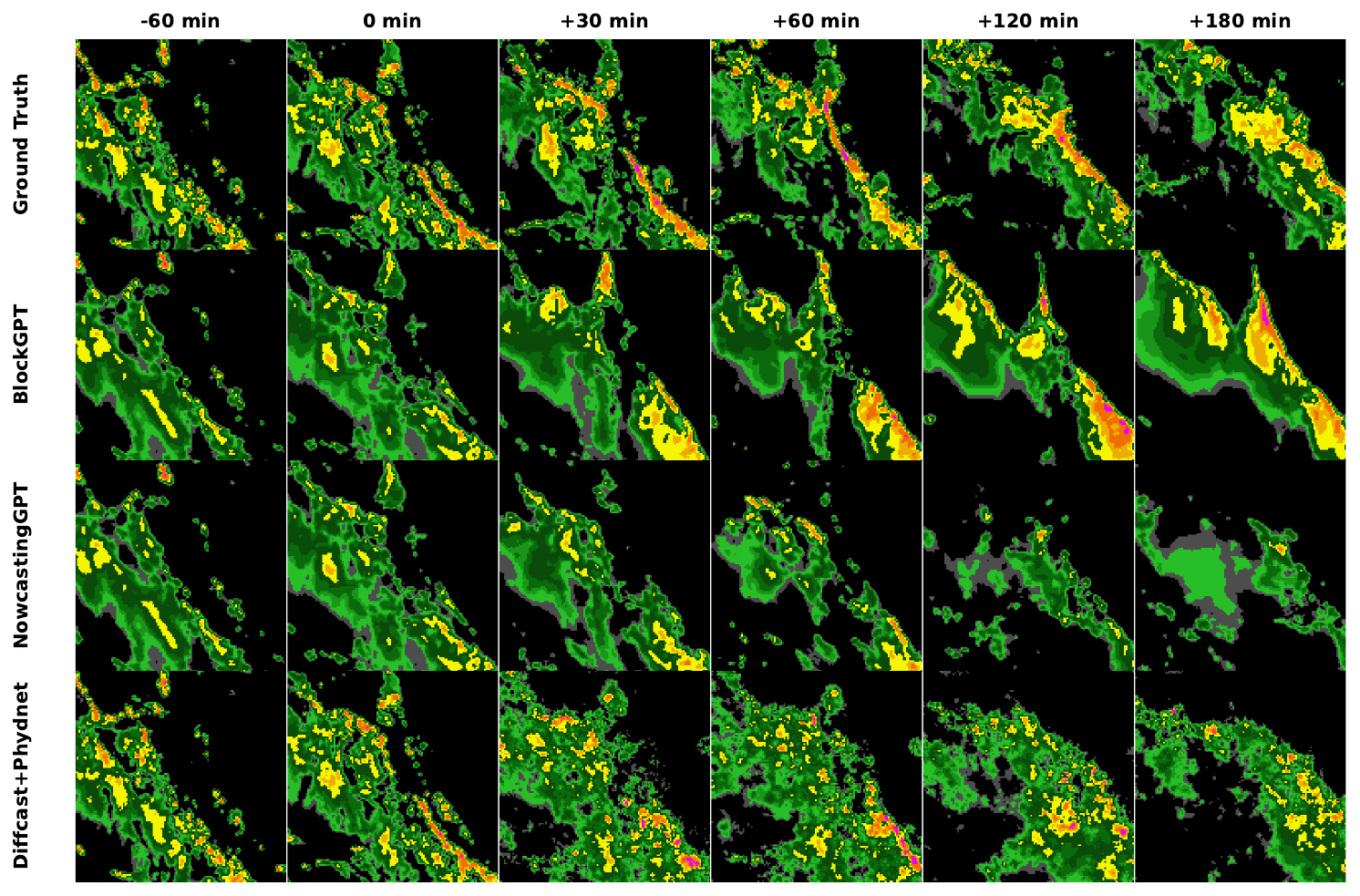}%
    }
    \caption{SEVIR Event~1 (linear system). BlockGPT best maintains structure and motion; a tendency to overestimate peak intensity emerges at longer lead times.}
    \label{fig:sevir_viz_exp1}
\end{figure}

\begin{figure}[h]
    \centering
    \makebox[\linewidth][l]{%
        \hspace*{-0.5cm}%
        \includegraphics[width=1.1\linewidth]{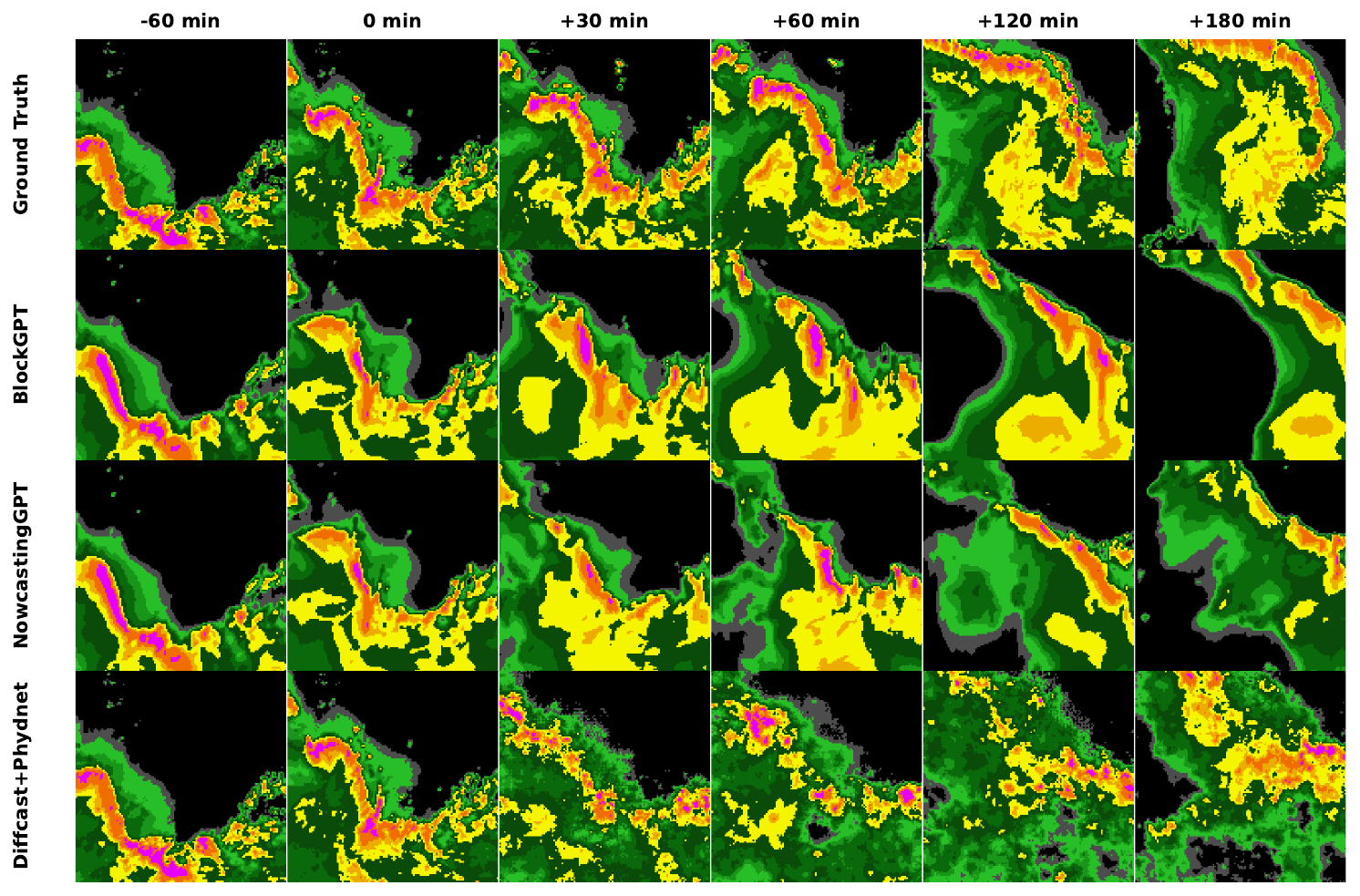}%
    }
    \caption{SEVIR Event~2 (circular organisation). BlockGPT best preserves the circular structure and its evolution; baselines lose shape and localisation.}
    \label{fig:sevir_viz_exp2}
\end{figure}

\paragraph{SEVIR.}
For the linear convective system in \autoref{fig:sevir_viz_exp1}, BlockGPT reproduces structure and propagation more faithfully than both baselines. At far lead times, it tends to overestimate peak intensity, consistent with the higher errors seen in the continuous metrics. In the circular/rotational case of \autoref{fig:sevir_viz_exp2}, BlockGPT again tracks geometry and location best; baselines lose the organised shape. Occasional overestimation of the high-intensity core at longer horizons aligns with our quantitative findings on SEVIR.

\subsection{Continuous Score Metrics across Percentile Levels}
\label{appendix:continuous-score-percentile-levels}
We evaluate continuous scores for aggregated data and across intensity-conditioned percentile bins (0--20, 20--40, 40--60, 60--80, 80--95th).

\textbf{KNMI.} At low-percentile bins, BlockGPT shows relatively higher MSE/MAE than the baselines; errors drop markedly toward higher-percentile bins where accurate prediction is operationally most critical. Across all bins, PCC is consistently higher for BlockGPT. Aggregated over all intensities, BlockGPT attains lower MSE/MAE and higher PCC than both benchmarks, indicating overall superiority. We also observe larger uncertainty across seeds for BlockGPT, attributable to the batched tokeniser design which can amplify seed-to-seed variability. Aggregated behaviour is summarised in \autoref{fig:metrics_all_levels_knmi30}.

\textbf{SEVIR.} Across bins, BlockGPT yields higher MSE/MAE than the baselines due to its strong sensitivity to high-intensity cores, which can be overestimated at longer lead times—incurring double-penalty effects from both intensity and displacement errors. Nevertheless, PCC remains competitive and typically exceeds DiffCast+Phydnet, indicating better spatial pattern fidelity despite larger amplitude errors. Aggregated trends are shown in \autoref{fig:metrics_all_levels_sevir30}.

\subsection{Catchment Analysis (KNMI only)}
\label{appendix:catchment-analysis}
\begin{figure}[h]
    \centering
    \makebox[\linewidth][l]{%
        \hspace*{-0.5cm}%
        \includegraphics[width=1.1\linewidth]{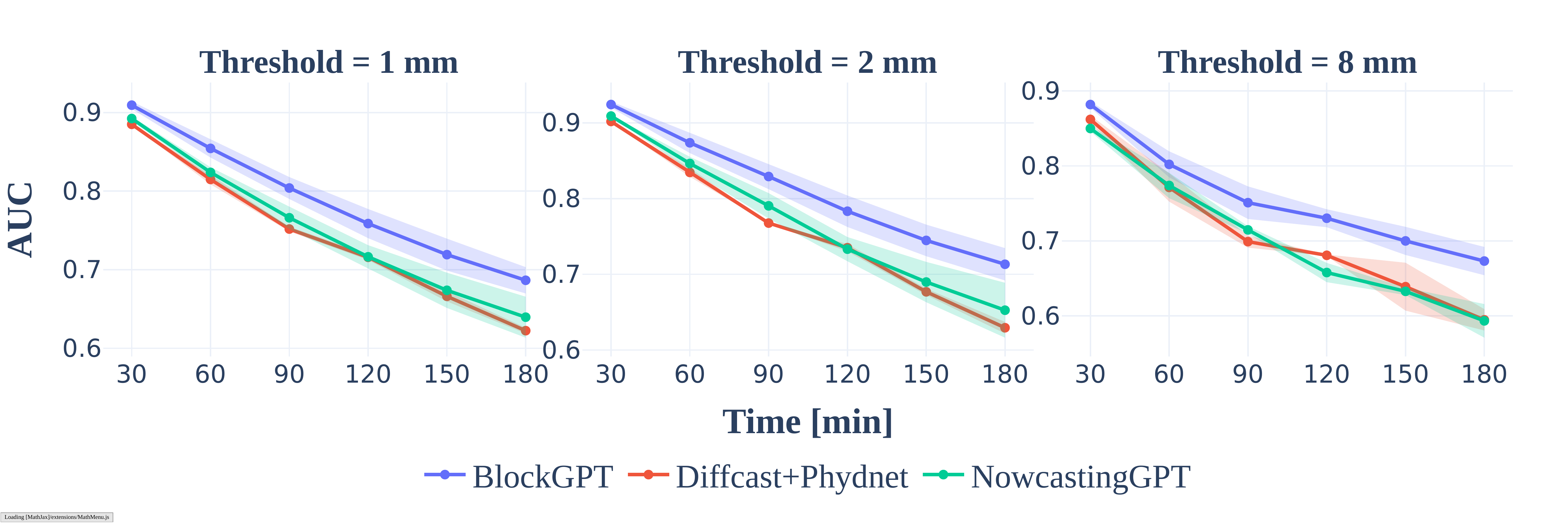}%
    }
    \caption{KNMI catchments: AUC-ROC over lead time for thresholds 1, 2, and 8~mm\,h$^{-1}$. Skill declines with lead time for all methods, but BlockGPT dominates across thresholds and horizons.}
    \label{fig:knmi_roc}
\end{figure}

\begin{figure}[h]
    \centering
    \makebox[\linewidth][l]{%
        \hspace*{-0.5cm}%
        \includegraphics[width=1.1\linewidth]{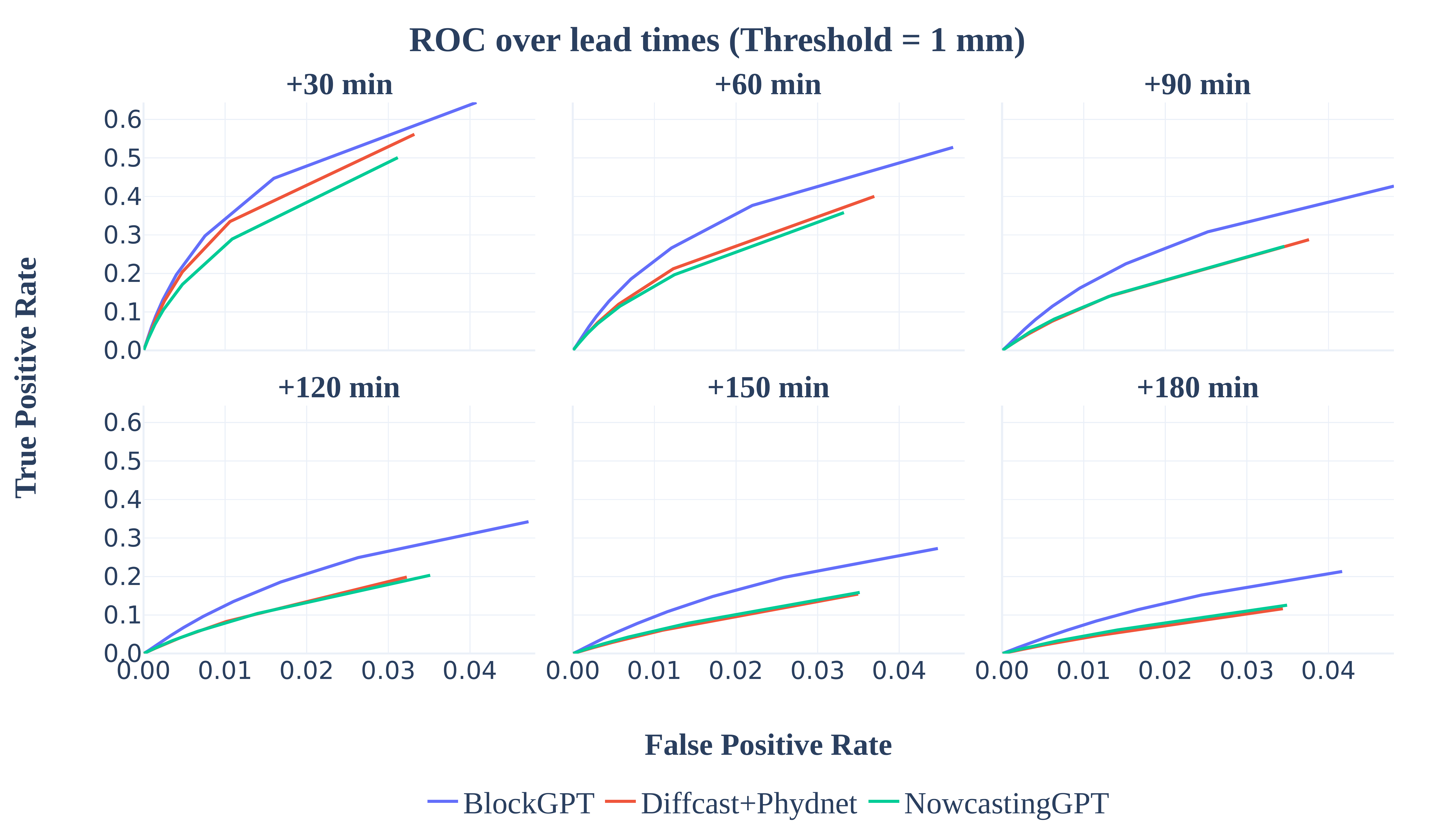}%
    }
    \caption{ROC curves across lead times for a 1\,mm\,h$^{-1}$ threshold on the KNMI (30\,min) dataset. Panels correspond to +30, +60, +90, +120, +150, and +180\,min lead times. Curves are averaged across seeds; performance improves as curves approach the top-left corner. Models compared: BlockGPT, DiffCast+Phydnet, and NowcastingGPT.}
    \label{fig:knmi_roc_1mm}
\end{figure}

\begin{figure}[h]
    \centering
    \makebox[\linewidth][l]{%
        \hspace*{-0.5cm}%
        \includegraphics[width=1.1\linewidth]{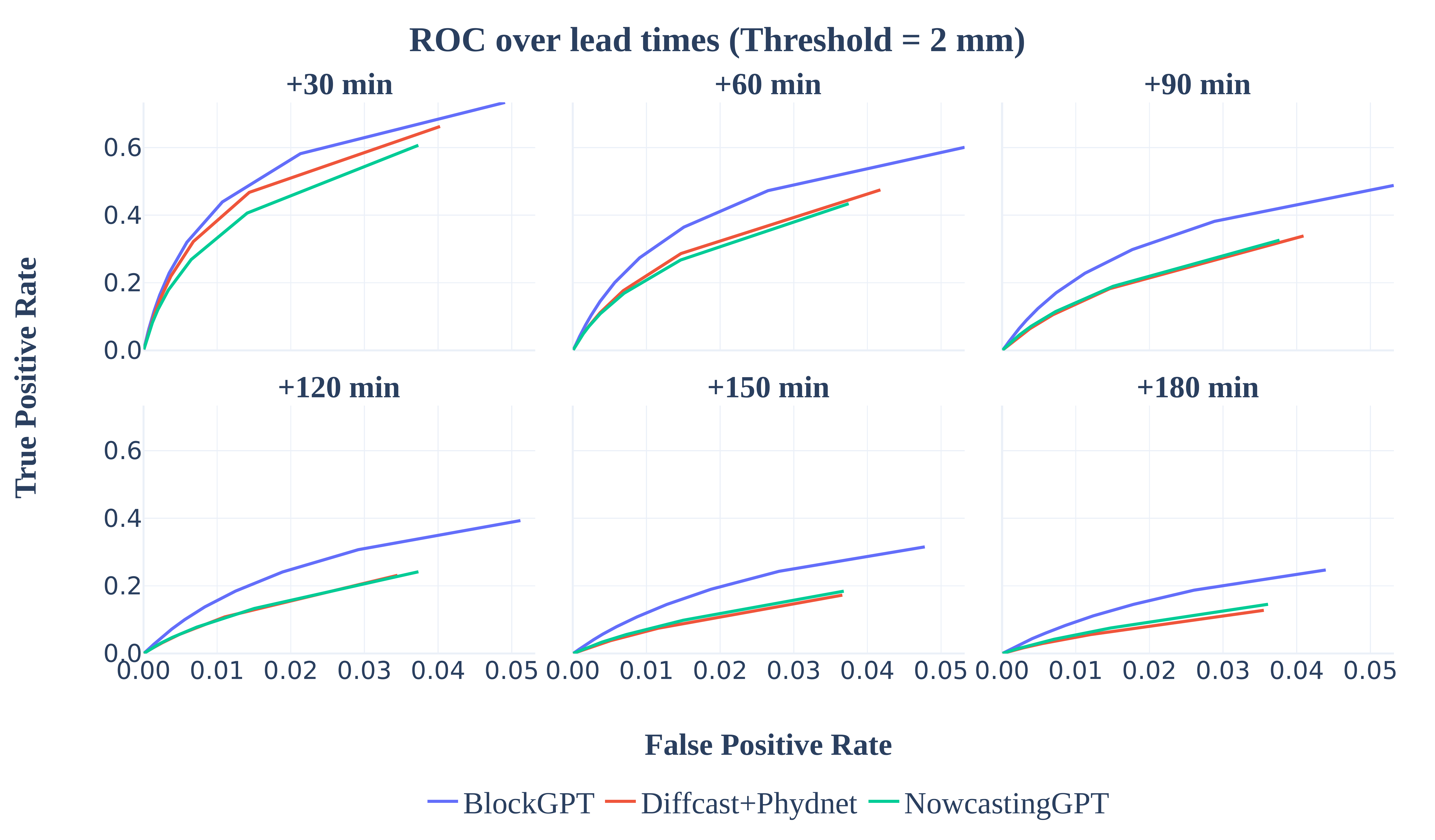}%
    }
    \caption{ROC curves across lead times for a 2\,mm\,h$^{-1}$ threshold on the KNMI (30\,min) dataset. Panels correspond to +30, +60, +90, +120, +150, and +180\,min lead times. Curves are averaged across seeds; performance improves as curves approach the top-left corner. Models compared: BlockGPT, DiffCast+Phydnet, and NowcastingGPT.}
    \label{fig:knmi_roc_2mm}
\end{figure}

\begin{figure}[h]
    \centering
    \makebox[\linewidth][l]{%
        \hspace*{-0.5cm}%
        \includegraphics[width=1.1\linewidth]{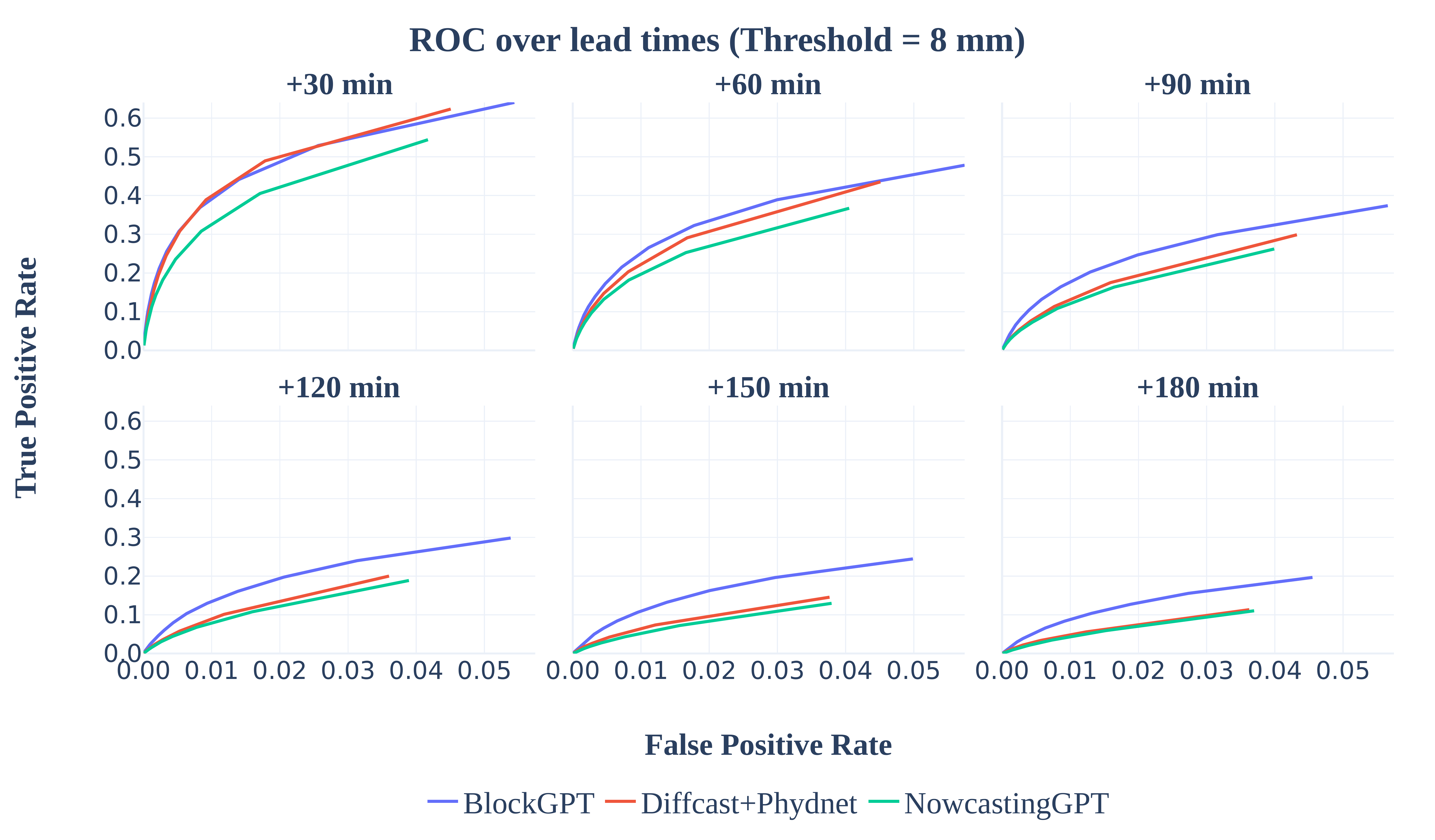}%
    }
    \caption{ROC curves across lead times for a 8\,mm\,h$^{-1}$ threshold on the KNMI (30\,min) dataset. Panels correspond to +30, +60, +90, +120, +150, and +180\,min lead times. Curves are averaged across seeds; performance improves as curves approach the top-left corner. Models compared: BlockGPT, DiffCast+Phydnet, and NowcastingGPT.}
    \label{fig:knmi_roc_8mm}
\end{figure}

We assess event-detection skill over hydrologically critical KNMI subregions \cite{imhoff2020spatial} using ROC and AUC-ROC at thresholds 1, 2, and 8~mm\,h$^{-1}$ for lead times 30, 60, 90, 120, 150, and 180\,min. ROC curves are computed for each threshold; AUC summarises skill across false-positive rates. As shown in \autoref{fig:knmi_roc}, BlockGPT consistently outperforms both NowcastingGPT and DiffCast+Phydnet in AUC-ROC at all thresholds and lead times, evidencing robustness across intensity and horizon. As expected, detection skill diminishes with increasing lead time for all models, yet BlockGPT maintains the best performance across subregions. The ROC curves at a 1, 2 and 8 mm$h^{-1}$ thresholds, stratified by lead time, are provided in \autoref{fig:knmi_roc_1mm}, \autoref{fig:knmi_roc_2mm} and \autoref{fig:knmi_roc_8mm}.

